%% file: main.tex
\documentclass[letterpaper]{article} 
\usepackage{aaai24}  
\usepackage{times}  
\usepackage{helvet}  
\usepackage{courier}  
\usepackage[hyphens]{url}  
\usepackage{graphicx} 
\def\UrlFont{\rm}  
\usepackage{natbib}  
\usepackage{caption} 
\usepackage{amsmath}
\usepackage{amssymb}
\usepackage{kotex}
\usepackage{verbatim}
\usepackage{multirow}
\usepackage{subcaption}
\usepackage{booktabs}
\usepackage{pifont}
\newcommand{\cmark}{\ding{51}}

\usepackage{algorithm}
\usepackage{algorithmic}

\usepackage{newfloat}
\usepackage{listings}
\DeclareCaptionStyle{ruled}{labelfont=normalfont,labelsep=colon,strut=off} 
\floatstyle{ruled}
\newfloat{listing}{tb}{lst}{}
\floatname{listing}{Listing}

\title{Expand-and-Quantize: Unsupervised Semantic Segmentation\\Using High-Dimensional Space and Product Quantization}
\author{
    Jiyoung Kim, Kyuhong Shim, Insu Lee, Byonghyo Shim \\
}
\affiliations{
    Department of Electrical and Computer Engineering, Seoul National University, Korea \\

    \{jykim, khshim, islee, bshim\}@islab.snu.ac.kr
}

\begin{document}
\maketitle

\input{section/0_abstract.tex}
\input{section/1_introduction}
\input{section/2_relatedwork}
\input{section/3_method}
\input{section/4_experiment}

\input{section/5_analysis}
\input{section/6_conclusion}
\input{section/7_appendix}
\newpage
\input{section/8_acknowledgment}
\bibliography{aaai24}

\end{document}

%% file: section/0_abstract.tex
\begin{abstract}
Unsupervised semantic segmentation (USS) aims to discover and recognize meaningful categories without any labels. 
For a successful USS, two key abilities are required: 1) information compression and 2) clustering capability.
Previous methods have relied on feature dimension reduction for information compression, however, this approach may hinder the process of clustering.
In this paper, we propose a novel USS framework called Expand-and-Quantize Unsupervised Semantic Segmentation (EQUSS), which combines the benefits of high-dimensional spaces for better clustering and product quantization for effective information compression.
Our extensive experiments demonstrate that EQUSS achieves state-of-the-art results on three standard benchmarks.
In addition, we analyze the entropy of USS features, which is the first step towards understanding USS from the perspective of information theory.
\end{abstract}

%% file: section/1_introduction.tex
\section{Introduction}\label{sec:intro}

Semantic segmentation, a task to classify every pixel to the proper category, has numerous applications including autonomous driving~\cite{auto1,auto2}, scene understanding~\cite{scene1,scene2}, and medical diagnostics~\cite{med1,med2}. 
Despite its significance, dissemination of supervised semantic segmentation is a bit slow, in particular for real-world applications due to the difficulty and cost of the pixel-wise annotations.
As a surrogate, unsupervised semantic segmentation (USS) has received much attention recently since it can save the time and effort to label the dataset~\cite{HP,STEGO,picie,IIC}.
\begin{figure}
    \vspace{-0.7cm}
    \centering
    \includegraphics[width=1\linewidth]{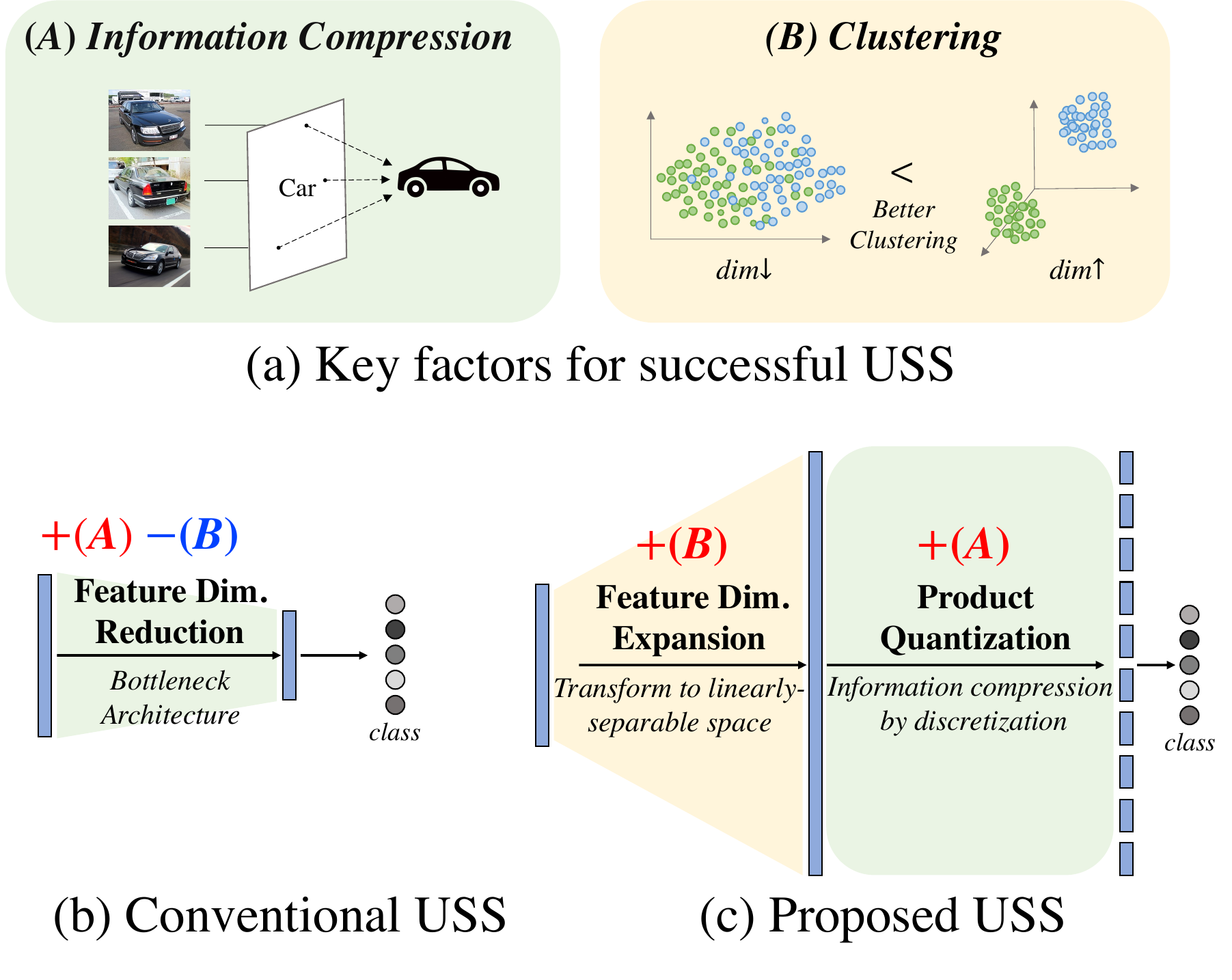}
    \caption{
    To achieve success in USS, two key factors should be considered: (\textit{A}) Information Compression and (\textit{B}) Clustering.
    The conventional model typically employs feature dimension reduction for (\textit{A}), but it can be counterproductive for (\textit{B}).
    In contrast, our proposed model takes advantage of both by employing feature dimension expansion for (\textit{B}) and product quantization for (\textit{A}).
    }
    \label{fig:concept}
    \vspace{-0.5cm}
\end{figure}

In most USS models, a pre-trained `backbone' module is employed to extract rich features from the input image.
The output of the backbone is a high-dimensional feature vector containing both class-relevant and class-irrelevant information~\cite{IIC,STEGO}.
In the absence of meaningful guidance, a USS model cannot efficiently judge which pixels are semantically similar.
To handle the issue, a USS `head' module should refine the feature such that the class-relevant information is preserved while class-irrelevant information is discarded.
In the information-theoretic perspective, this process can be considered as a form of \textit{lossy information compression}~\cite{lossy}.
For instance, if there are 27 classes~\cite{COCO}, a perfect classification can be achieved with 5 bits, which is far smaller than the amount of information contained in the 384-dim feature vector (384 elements $\times$ 32-bit floating point = 12,288 bits).
After the information compression, USS model clusters the outputs using k-means such that pixels in the same group are aggregated to the same class.

So far, the primary goal of USS frameworks is to compress information while preserving as much class-relevant information as possible.
Popularly used method to achieve the goal is \textit{feature dimension reduction} (e.g., 384-dim to 70-dim) for information compression, followed by training to retain class-relevant information in the compressed feature~\cite{AC,IIC,picie,transFGU,STEGO} (see Fig.~\ref{fig:concept}(b)).
Despite its design simplicity and low-complexity, conventional USS framework has a serious shortcoming; loss of class-relevant information and introduction of quantization noise.
Without proper guidance, features cannot be \textit{easily clustered} in a metric space.
The moral of the story is that the features should achieve a higher level of `clusterability'.
We henceforth use this term as the capability to achieve accurate decision boundaries between classes.

To facilitate the clustering of features (i.e., better clusterability), we cast the feature vector into a high-dimensional space.
This approach is motivated by the well-known fact that a nonlinear transformation to a high-dimensional latent space can enhance the model's ability to find hyperplanes that properly separate classes~\cite{kernel}.
Essentially, as the dimension increases, the volume of the space where the feature vector exists also increases exponentially.
This makes the class-relevant information more sparsely distributed, facilitating the model easier to identify linear decision boundaries between groups of data (classes).
As a result, training methods for enhancing the class-relevance of features can be more effective in high-dimensional space than in low-dimensional space.

Since previous USS approaches adopt feature dimension reduction architecture, they could not achieve both the benefits of high-dimensional space and information compression at the same time (see Fig.~\ref{fig:concept}(b)).
In contrast, we aggressively exploit \textit{feature dimension expansion} to improve clusterability for USS.
To make the most of the benefits of high-dimensional space while achieving information compression, we exploit the \textit{product quantization} (see Fig.~\ref{fig:concept}(c)).
Intuitively, this quantization process can be interpreted as a filtering mechanism that preserves only the common information represented by cluster centroids and discards unwanted class-irrelevant information and the quantization noise.
We expect that the centroids represent the class-relevant information, as long as the features are properly clustered.

In this paper, we propose an entirely different USS framework referred to as Expand-and-Quantize Unsupervised Semantic Segmentation (\textbf{EQUSS}).
In order to take advantage of both high-dimensional space for improved clustering and PQ for class-aware information compression, main operation of EQUSS is divided into two steps: `expand' and `quantize'.  
To the best of our best knowledge, we are the first to employ either high-dimensional features or PQ in previous USS literature.

By measuring the number of bits to represent a class, we show that EQUSS uses significantly smaller number of bits than the previous state-of-the-art (SOTA) method~\cite{STEGO} (246 bits vs. 475 bits).
We also show that EQUSS allocates bits to each class according to the representation difficulty of the class.
For example, EQUSS allocates fewer bits for classes with small variations (i.e., sky, water) while more bits for classes with divergent appearances (i.e., sports, indoor).
We find that this behavior is a unique characteristic of EQUSS and is not observed in other USS models.
From our empirical experiments on CocoStuff-27~\cite{COCO}, Cityscapes~\cite{cityscapes}, and Potsdam-3 USS benchmarks, we show that the proposed EQUSS outperforms the recent SOTA~\cite{STEGO} by a substantial margin.

The contributions of our work are as follows:
\begin{itemize}

    \item We propose a novel USS framework referred to as EQUSS, that utilizes high-dimensional spaces to improve clusterability. 
    We are the first to recognize the critical role of clusterability in USS architecture design. 
    
    \item We exploit the product quantization for the information compression instead of widely used feature dimension reduction.
    From our experiments, we show that EQUSS achieves SOTA performance on three USS benchmarks in all metrics.

    \item We quantify the information capacity of USS features in terms of \textit{bits}. 
    Our analyses establish a relationship between the entropy of features and segmentation accuracy, providing valuable insights guideline for USS design.
    
\end{itemize}

%% file: section/2_relatedwork.tex
\section{Related Work}\label{sec:related_work}

\subsection{Unsupervised Semantic Segmentation}~\label{ssec:uss}
Over the years, various training methods for USS have been proposed to retain class-relevant information while removing irrelevant ones.
Due to the absence of labels, previous USS approaches focused on developing reliable guidance that is presumably related to the class.
One line of research is the mutual information-based learning~\cite{IIC, AC}.
The main idea behind this approach is that each pixel's class remains the same after image augmentations.
For example, IIC~\cite{IIC} maximizes the mutual information between different views of the inputs generated by the augmentation.
Similarly, PiCIE~\cite{picie} exploits multiple photometric and geometric transformations to learn consistent class representations.
Another line of research is the k-means clustering-based methods~\cite{DeepCluster,picie,transFGU} that generate pixel-level pseudo-labels to train the model with a classification loss.
TransFGU~\cite{transFGU} obtains high-level semantic information from feature clustering and generates pseudo-labels from Grad-CAM~\cite{grad_cam}.
These pseudo-labels can be refined iteratively by altering between clustering and training steps~\cite{picie}.
Recently, STEGO~\cite{STEGO} suggests the feature correspondence distillation from backbone features to head features.
A common ingredient in the aforementioned studies is to use feature dimension reduction as a means of information compression and then apply the training techniques to the compressed features.
In contrast, in our EQUSS, we expand the feature dimension and train these expanded features to contain class-relevant information.

\subsection{Information Bottleneck}~\label{ssec:ib}
Feature dimension reduction has been widely used in many fields of designing information bottleneck architecture~\cite{bottleneck,tishby}.
This architecture has proven to be useful in a wide range of applications, including image classification~\cite{resnet}, latent representation learning~\cite{vae}, model compression~\cite{lowrank_compress}, and lightweight model adaptation~\cite{lowrank_lora}.
These applications, except for USS, typically have clear guidance for compression.
For example, in the image classification, the model learns to identify classes using the supervised learning.
In the latent representation learning, a model aims to find a compact representation of an input by minimizing the reconstruction error.
As discussed, USS lacks such clear guidance so the feature dimension reduction might not be a proper choice.
Instead of relying on the feature dimension reduction, our EQUSS exploits the product quantization for information bottleneck design.

\subsection{Product Quantization}~\label{ssec:pq}
Product Quantization (PQ) was initially proposed for the nearest neighbor search~\cite{pq}.
Due to its ability to compress high-dimensional vectors with extremely fast searching, PQ has been popularly used in large-scale retrieval tasks~\cite{pq_retrieval,SPQ,dpq}.
For example, SPQ~\cite{SPQ} has employed PQ for the unsupervised image retrieval by jointly learning codewords and feature extractors in a self-supervised manner.
Recently, PQ has been applied to other domains such as neural network compression~\cite{pq_gnn,comp1,quantnoise}, text-to-speech synthesis~\cite{tts1,tts2}, and zero-shot learning~\cite{PQZSL}.
While these studies mainly utilize PQ's ability for fast indexing and data size reduction, we exploit PQ in the information compression.
To the best of our knowledge, we are the first to utilize PQ for semantic segmentation.

%% file: section/3_method.tex
\section{Expand-and-Quantize Unsupervised\\Semantic Segmentation (EQUSS)}\label{sec:method}
\begin{figure}[t] %
    \centering   \includegraphics[width=1.0\linewidth]{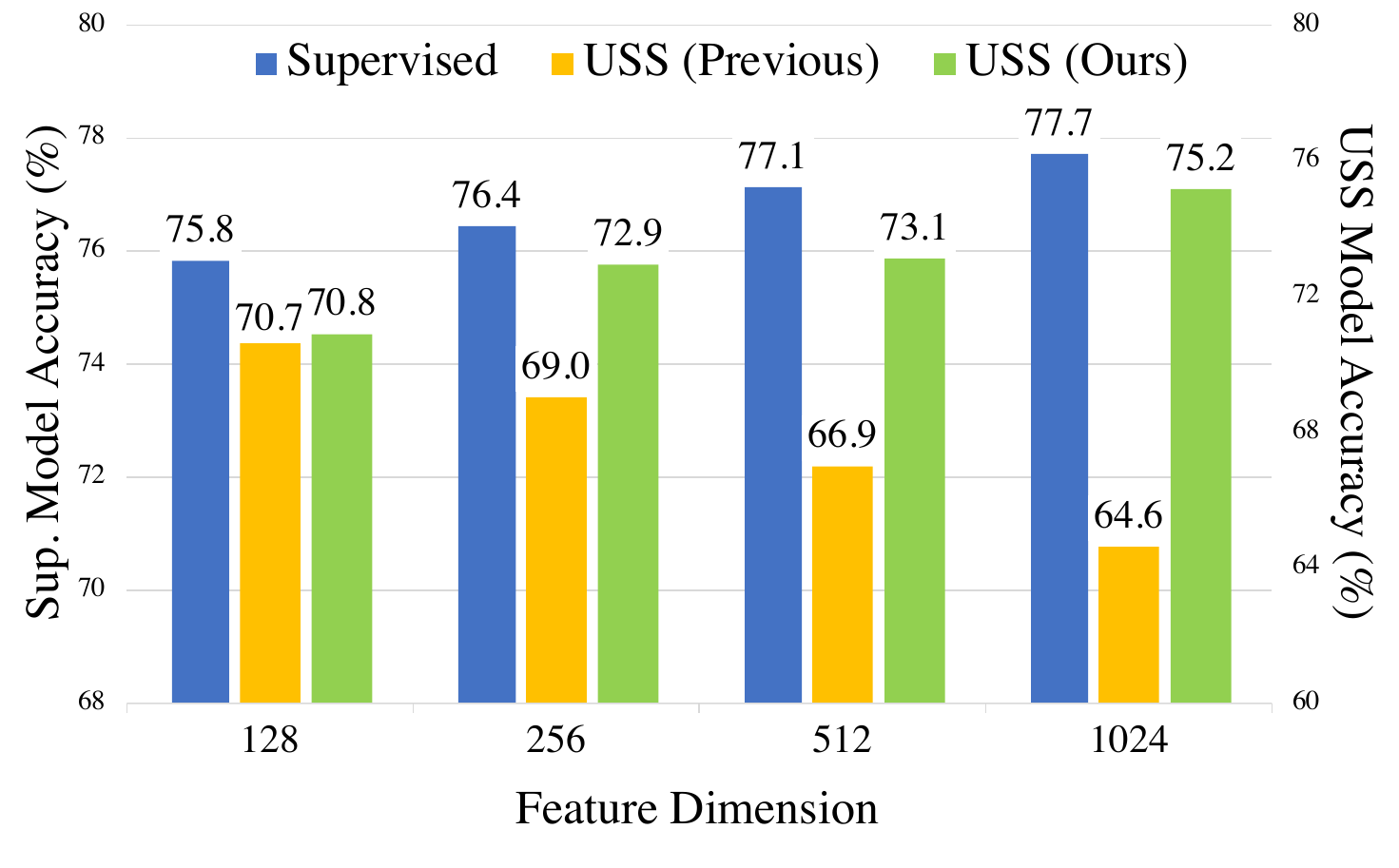}
    \vspace{-0.2cm}
    \caption{
    The relationship between feature dimension size and segmentation accuracy on the supervised model, previous USS model~\cite{STEGO}, and our proposed USS model.}
    \label{fig:sl_unsl}
    \vspace{-0.5cm}
\end{figure}
\begin{figure*}[t]
    \centering
    \includegraphics[width=1\linewidth]{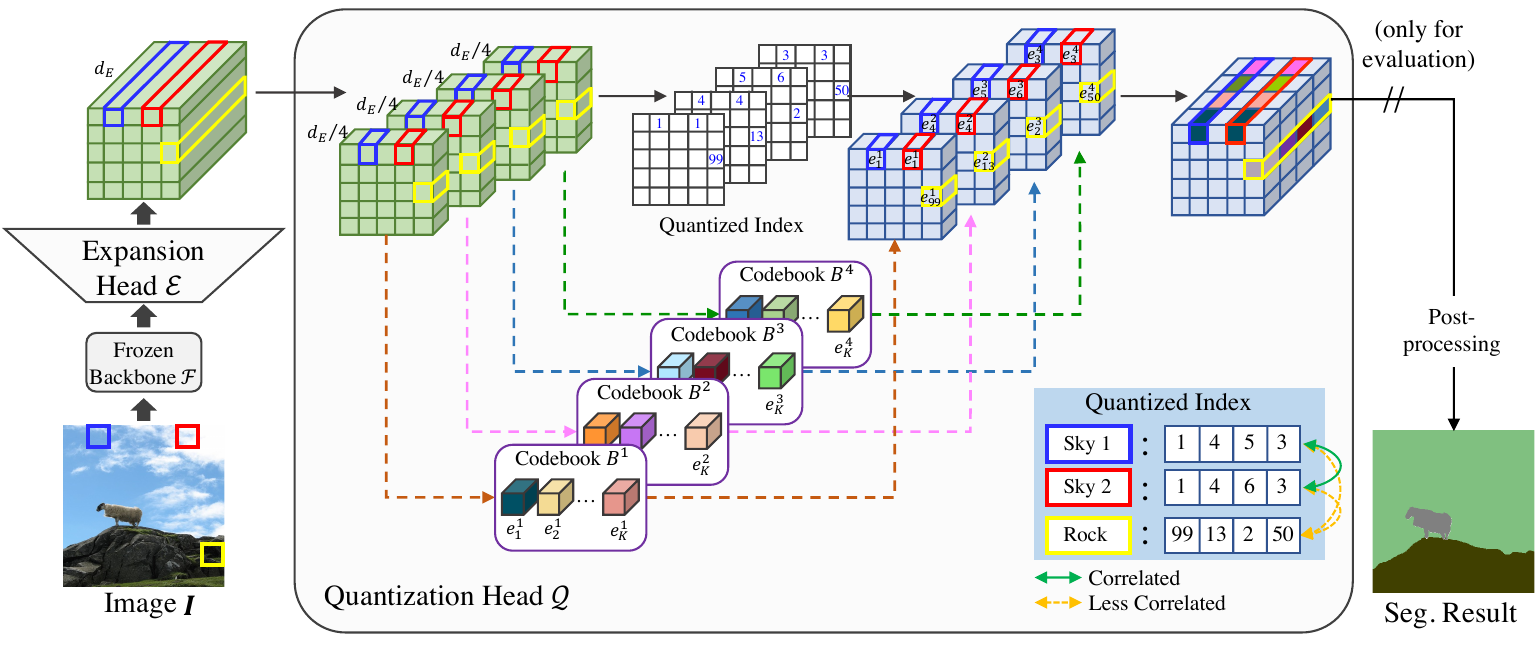}
    \caption{
    Overview of the proposed EQUSS.
    The expansion head $\mathcal{E}$ first expands the feature extracted from the backbone $\mathcal{F}$ into high-dimensional spaces.
    Then, the quantization head $\mathcal{Q}$ applies product quantization to generate the quantized output.
    Finally, this output is used for clustering and linear probing during the evaluation.
    For a better understanding, we set 
    $M=4$.
    }
    \label{fig:overall}
\end{figure*}

In this section, we first present motivating experiments describing our central idea and then discuss two key components of EQUSS: feature dimension expansion and product quantization.
Fig.~\ref{fig:overall} illustrates the overview of the proposed framework, EQUSS.

\subsection{Motivation}\label{sec:method:motivation}
In Fig.~\ref{fig:sl_unsl}, we plot the segmentation accuracy of the supervised and unsupervised semantic segmentation models for different feature dimensions.
We show that an increase in the feature dimension is beneficial for the supervised model but not for the previous USS model (see blue and yellow bars in Fig.~\ref{fig:sl_unsl}).

We believe that feature dimension expansion can help the supervised model accurately delineate the decision boundary between different classes. 
This can be supported by the celebrated Cover's theorem~\cite{cover_theorem}, saying that patterns (classes in our context) that are difficult to separate in low-dimensional space might be better separable when projected into high-dimensional space by the nonlinear transformation.
In other words, feature dimension expansion can contribute to better clusterability.
We note that such behavior cannot be contained by the conventional USS model since the dimension reduction significantly degrades the clusterability.
We claim that naively increasing the feature dimension may result in the loss of the ability to compress information.
Despite the potential benefits of high-dimensional features for clustering, the absence of information compression causes a detrimental performance loss.

Motivated by these findings, we raise a question: Can we achieve the best of both, better clusterability and information compression? 
In this paper, we propose a novel architecture that leverages the expansion of feature dimensions without sacrificing the capability to compress information.

\subsection{Feature Extraction}
Given an unlabeled dataset $\textit{D}=\{I_1, I_2, \ldots, I_N\}$ of \textit{N} training images, the goal of USS is to learn a model that maps $\textit{I}{(i,j)}\mapsto c\in C$, where (\textit{i}, \textit{j}) represents a spatial location in an image \textit{I} and \textit{C} indicates a set of classes.
To extract rich visual features, the backbone model $\mathcal{F}$ takes an image \textit{I} and returns the features \textit{F} $\in \mathbb{R}^{d_F\times h\times w}$ where $d_F$ and $(h,w)$ indicate the feature dimension and spatial size, respectively.
We denote a feature vector at $(i, j)$ as $f_{(i,j)} = F(i,j) \in \mathbb{R}^{d_F}$; for simplicity, we omit the spatial indices $(i, j)$ for $f$.

\subsection{Feature Dimension Expansion and Product Quantization}\label{ssec:method}
\paragraph{Expansion}
For the information refinement process discussed in Sec.~\ref{sec:intro}, the extracted feature $f$, obtained from the backbone, passes through an expansion head $\mathcal{E}$.
Essentially, the role of this head is to perform the nonlinear transform on $f$ to generate the expanded feature vector $x\in \mathbb{R}^{d_E}$ where ${d_E}$ is the dimension of the expanded feature.
Note that $d_E \ll d_F$ for the conventional USS framework and $d_E > d_F$ for our EQUSS framework.

\paragraph{Quantization}
In order to compress the information, the quantization head $\mathcal{Q}$ takes an expanded feature vector $x$ as an input and generates a quantized output $q \in \mathbb{R}^{d_E}$.
Note that the quantization process does not change the feature dimension.
In our study, we employ PQ instead of vector quantization (VQ).
The essence of PQ is to decompose a high-dimensional space into a Cartesian product of low-dimensional subspaces and then individually apply VQ for each partition.
In doing so, PQ can exponentially increase the pool of possible candidates without greatly increasing the number of codewords for each codebook.

Specifically, the quantization head $\mathcal{Q}$ consists of $M$ codebooks $\{ B^1, \ldots, B^M\}$, each of which contains $K$ codewords as $B^m =\{e^m_1, \ldots,e^m_K\}$ where $e^i_j \in \mathbb{R}^{d_E\slash M}$ is the \textit{j}-th codeword in the \textit{i}-th codebook $B^i$.
In the first stage of PQ, $x$ is divided into multiple subvectors $\{x^1, \ldots, x^M\}$ as
\begin{equation}
    x = \texttt{concatenate}[x^1, \ldots, x^M], \quad x^i \in \mathbb{R}^{d_E\slash M}
\end{equation}
For the second stage, each subvector $x^m$ is mapped to the closest codeword in the \textit{m}-th codebook as:
\begin{align}
    q^m &= Q^m(x^m) = e^m_k , \nonumber \\
    \text{where } k &= \arg\min_j\Vert \hat{x}^m - \hat{e}^m_j\Vert^2_2.  \label{eq:argmin}
\end{align}
$Q^m$ is a quantization operator for \textit{m}-th subvector.
The distance function is a squared Euclidean distance between $\hat{x}^i$ (= ${x^i} / \Vert {x^i} \Vert_2$) and $\hat{e}^i_j$ (= ${e^i_j} / \Vert {e^i_j} \Vert_2$).
We empirically discover that this normalization is essential for stabilizing the training as we initialize codewords from a random distribution.
Finally, by concatenating $M$ quantized outputs $\{q^1, \ldots, q^M\}$, we obtain the final output $q = \texttt{concatenate}[q^1, \ldots, q^M]$.

\subsection{Training Objective}
The loss for PQ consists of two terms: codebook loss and commitment loss.
The codebook loss minimizes the distance between each subvector and its selected codeword.
Then, the commitment loss encourages the subvector to stay close to the chosen codeword.
The codebook loss and the commitment loss are formally defined as:
\begin{align}
    L_{codebook} &= \frac{1}{M}\sum_{m=1}^{M}\Vert{sg[x^m] - e^m_k}\Vert_2^2 \label{eq:book} \\
    L_{commit} &= \frac{1}{M}\sum_{m=1}^{M}\Vert{x^m - sg[e^m_k]} \Vert_2^2 \label{eq:commit}
\end{align}
where $sg$ denotes the stop-gradient operation. 
Since the operation $\arg\min$ in Eq.~\eqref{eq:argmin} is non-differentiable, we approximate the gradient of $x^i$ by copying gradients from a selected codeword $e^i_k$.
Note that this process is known as the straight-through gradient estimation technique~\cite{gradient_estimator}.

The overall objective function is the sum of the training loss for the expansion head $L_{head}$~\cite{STEGO}, codebook loss $L_{codebook}$, and commitment loss $L_{commit}$:
\begin{align}
    L &= L_{head} + \lambda_1 L_{codebook} + \lambda_2 L_{commit}
\label{eq:total_loss}
\end{align}
where $\lambda_1$, $\lambda_2$ $>$ 0 are weighting coefficients.
Please refer to previous work~\cite{STEGO} for more details about $L_{head}$.

\subsection{Other Details}\label{sec:eval}
\paragraph{Evaluation Protocol}
For measuring the performance of the USS model, we need additional steps that map the learned feature $q$ to the ground truth label $c$.
The USS performance is evaluated with two measurement processes, namely, linear probing and unsupervised clustering. 
Note that these measuring processes do not update the model parameters as they are completely separated from USS training (see Fig.~\ref{fig:overall}).
In Appendix, we discuss more details on our evaluation process.

\paragraph{CRF}
Using Conditional Random Field (CRF)~\cite{crf}, a widely used post-processing in semantic segmentation, one can optionally refine the final segmentation mask.
CRF is a probabilistic graphical model that utilizes the relationship between neighboring pixels, such as their distances or RGB values.
CRF improves the result by sharpening the edges and reducing the noises based on statistical relationships between pixels. 

\subsection{Measuring the Information Capacity}\label{sec:method_entropy}
To quantify the ability for preserving class-relevant information after the information compression, we measure the Shannon entropy~\cite{shannon_entropy}.
Intuitively, an increase in class-irrelevant information will cause an increase in the number of bits to represent the feature.

Since the mechanism of EQUSS is based on PQ, we can easily obtain an empirical probability mass function (PMF) over codewords in each codebook based on the frequency.
For the given set of features $X$, we first calculate $p_i^m$, a probability that a subvector $x^m$ is mapped to the $e_i^m$ as
\begin{align}
    p_i^m &= \mathbb{E}_X[P(q^m=e_i^m)] \nonumber \\
    &\approx \frac{1}{|X|} \sum_{x\in X} \mathbb{I}(Q^m(x^m)=e_i^m)
\end{align}
where $\mathbb{I}(\cdot)$ is an indicator function that returns 1 if the condition is true and 0 otherwise.
In practice, we approximate the expectation using a sufficient number of randomly selected features.
To quantify the amount of information on features, we define the sum of entropy $J(X)$ as
\begin{equation}
    J{(X)} = \sum_{m=1}^{M} \bigg(-\sum_{i=1}^K p_i^m \log_2(p_i^m) \bigg). 
    \label{eq:EQUSS_entropy}
\end{equation}
$J(X)$ can be interpreted as the number of bits to represent the entire feature.
In Sec.~\ref{sec:ab:entropy}, we compare the per-class entropy of EQUSS and STEGO~\cite{STEGO} for information capacity in representing each class.

%% file: section/4_experiment.tex
\section{Experiments}\label{sec:experi}
We evaluate EQUSS on three standard semantic segmentation datasets and compare with the recent SOTA methods. 
See Appendix for more information about datasets and implementation details.

\input{table/cocostuff.tex}
\subsection{USS Performance}
Table~\ref{tab:cocostuff} compares the performance of EQUSS with the recent USS models on CocoStuff-27.
EQUSS outperforms previous works by a large margin in all metrics.
In particular, EQUSS outperforms STEGO~\cite{STEGO} by \textbf{+5.5} in the unsupervised accuracy and \textbf{+2.9} in the linear mIoU, respectively.
Table~\ref{tab:sub_cityscapes} and \ref{tab:sub_potsdam} report the USS performance on Cityscapes and Potsdam-3, respectively.
While STEGO and TransFGU~\cite{transFGU} suffer from the trade-off between accuracy and mIoU on Cityscapes, EQUSS shows outstanding performances in both accuracy and mIoU.
In Potsdam-3, EQUSS outperforms STEGO by \textbf{+5.0} in the unsupervised accuracy.
We attribute this performance gain to 1) better clusterability of high-dimensional space and 2) effective compression of the class-irrelevant information.

\subsection{Qualitative Result}
In Fig.~\ref{fig:comparison}, we present the segmentation results of EQUSS and STEGO on CocoStuff-27.
We observe that STEGO is often biased to certain visual information such as colors or edges, which causes the mask of an object to be split into multiple parts.
For example, in Fig.~\ref{fig:comparison} (a) and (h), the written texts on the wall and rock behind a giraffe are inaccurately predicted.
In contrast, EQUSS accurately groups the pixels with diverse characteristics into the same class.
This is because EQUSS can effectively compress class-irrelevant information and output features that are highly correlated with classes.
Please see more qualitative comparisons in Appendix.
\input{table/sub_table}

%% file: table/cocostuff.tex
\begin{table}[t]
    \centering
    \resizebox{1.0\linewidth}{!}{
    \begin{tabular}{lllll}
    \toprule
    \multirow{2}{*}{Model} & \multicolumn{2}{c}{Unsupervised} & \multicolumn{2}{c}{Linear Probe} \\
    & Acc. & mIoU & Acc. & mIoU  \\
        \midrule
        ResNet50~\cite{resnet} & 24.6 & 8.9 & 41.3 &  10.2 \\
        MoCoV2~\cite{MoCoV2} & 25.2 & 10.4 & 44.4 &  13.2 \\
        DINO~\cite{DINO} & 30.5 & 9.6 & 66.8 &  29.4 \\
        Deep Cluster~\cite{DeepCluster}  & 19.9 & - & - & - \\
        SIFT~\cite{SIFT} & 20.2 & - & - &  - \\ 
        Isola et al.~\cite{isola2015} & 24.3 & - & - & - \\ 
        AC~\cite{AC} & 30.8 & - & - &  -\\ 
        IIC~\cite{IIC} & 21.8 & 6.7 & 44.5 & 8.4  \\ 
        MDC~\cite{picie} & 32.2 & 9.8 & 48.6 & 13.3 \\ 
        PiCIE~\cite{picie} & 48.1 & 13.8 & 54.2 & 13.9 \\ 
        PiCIE+H~\cite{picie}& 50.0 & 14.4 & 54.8 & 14.8 \\ 
        ACSeg~\cite{acseg}& - & 16.4 & - & - \\
        TransFGU (ViT-S)~\cite{transFGU} & 52.7 & 17.5 & - & - \\
        STEGO (ViT-S)~\cite{STEGO} & 48.3 & 24.5 & 74.4 & 38.3 \\
        \midrule
        \textbf{EQUSS} (ViT-S) & \textbf{53.8} & \textbf{25.8} & \textbf{75.2} & \textbf{41.2} \\
        \bottomrule
    \end{tabular}
    }
    \caption{USS performance comparison on \textbf{CocoStuff-27}. 
    }
    \label{tab:cocostuff}
    \vspace{-0.3cm}
\end{table}

%% file: table/sub_table.tex
\begin{table}[t]
        \begin{subtable}[h]{0.25\textwidth}
	\centering
    \resizebox{1\linewidth}{!}{
    \small
    \begin{tabular}{lcccc}
    \toprule
    \multirow{2}{*}{Model} & \multicolumn{2}{c}{Unsupervised} \\
     & Acc. & mIoU \\
    \midrule
    IIC & 47.9 &  6.4 \\
    MDC  & 40.7 &  7.1 \\
    PiCIE & 65.5 & 12.3 \\
    TransFGU (ViT-S) & 77.9 & 16.8 \\
    STEGO (ViT-B) & 73.2 & 21.0\\ 
    \midrule
    \textbf{EQUSS} (ViT-B) & \textbf{79.9} & \textbf{22.0} \\
    \bottomrule
    \end{tabular}
    }
    \caption{Cityscapes}
    \label{tab:sub_cityscapes}
	\end{subtable}
	\hfill
	\begin{subtable}[h]{0.2\textwidth}
    \centering
    \resizebox{1\linewidth}{!}{
    \small
    \begin{tabular}{lc}
    \toprule
        Model & U.Acc. \\ 
        \midrule
        Deep Cluster & 41.7 \\
        SIFT & 38.2\\ 
        Doersch et al. & 49.6\\ 
        Isola et al. & 63.9\\ 
        IIC & 65.1 \\ 
        STEGO (ViT-B) & 77.0\\
        \midrule
        \textbf{EQUSS}(ViT-B) & \textbf{82.0} \\
        \bottomrule
    \end{tabular}
    }
    \caption{Potsdam-3}
    \label{tab:sub_potsdam}
	\end{subtable}
	\caption{Performance comparisons on (a) \textbf{Cityscapes} and (b) \textbf{Potsdam-3}.
 }
 \label{tab:cityscape_potsdam}
 \vspace{-0.3cm}
\end{table}

%% file: section/5_analysis.tex
\section{Analysis}\label{sec:ablation}
\begin{figure*}[!t]
    \centering
    \includegraphics[width=0.95\textwidth]{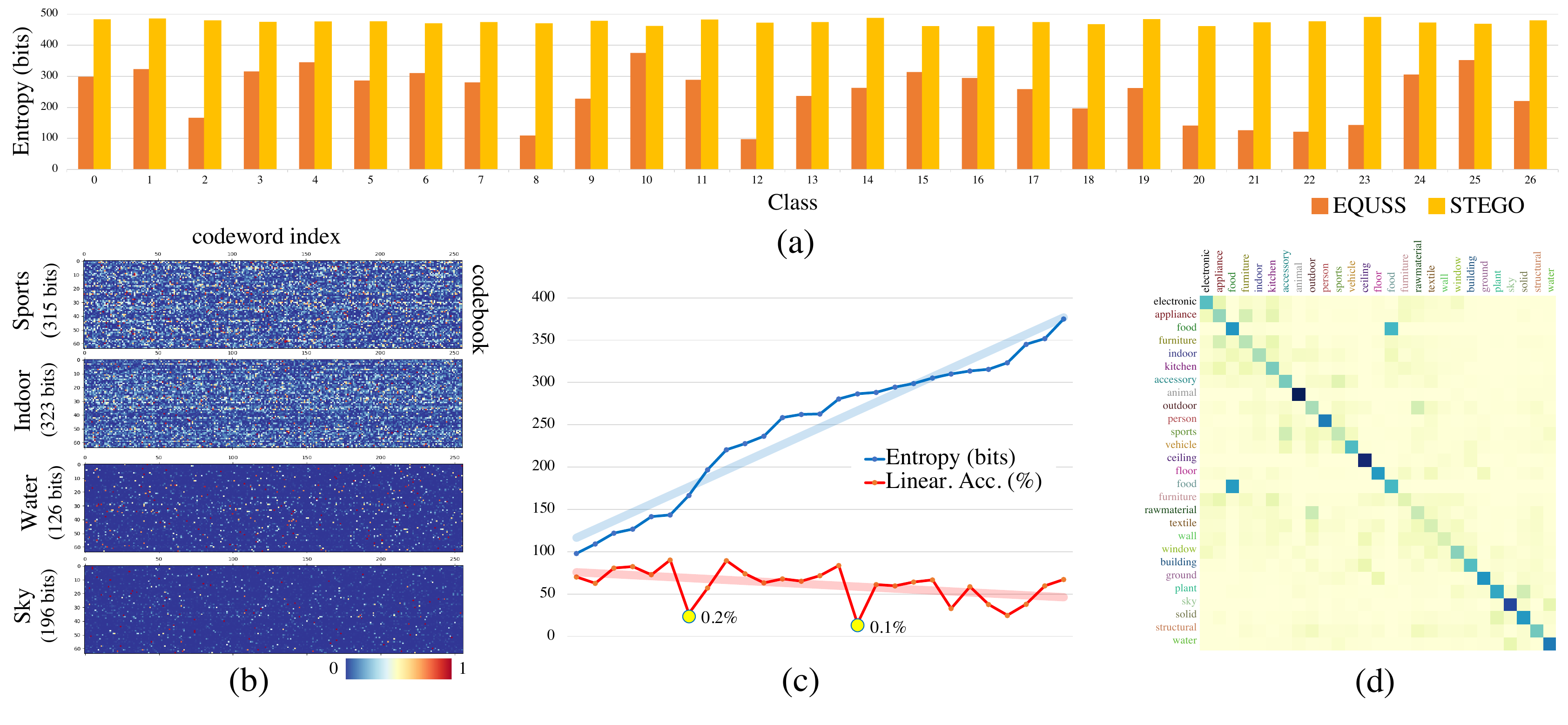}
    \vspace{-0.5cm}
    \caption{ 
    Analyses on EQUSS.
    (a) Per-class entropy of EQUSS and STEGO~\cite{STEGO}.
    (b) Frequency of each codeword to be selected, along with its corresponding codebook.
    (c) Relationship between entropy and accuracy. 
    Yellow points with particularly low accuracy correspond to the most infrequent minority classes (0.1\% and 0.2\%).
    (d) Inter-class distance between the combination of codewords. 
    The darker the color, the closer the distance.
    Best viewed in PDF.
    }
    \label{fig:analysis}
    \vspace{-0.4cm}
\end{figure*}
\subsection{Relationship between Entropy and Accuracy}\label{sec:ab:entropy}
As discussed in Sec.~\ref{sec:method_entropy}, we quantify the information capacity of features in terms of bits.
Specifically, we compare the per-class entropy of EQUSS and STEGO on the CocoStuff-27 (see Fig.~\ref{fig:analysis}(a)).
Unlike EQUSS, STEGO outputs a continuous variable, so we cannot directly measure the entropy using Eq.~\eqref{eq:EQUSS_entropy}.
To deal with the problem, we discretize the continuous distribution by histogram and calculate entropy for each feature dimension based on the frequency (see Appendix).

The most noticeable observation is that the entropy of EQUSS is far smaller than that of STEGO, 246 bits vs. 475 bits, on average.
Intuitively, the number of bits corresponds to the number of distinct representations for each class.
If there are few combinations of codewords for representing a certain class, we can assume that such candidates contain common core information shared among the samples in that class.
Consequently, when the class is represented with fewer candidates, the output features are also more likely to contain class-relevant information, empirically supported by our superior USS performance.

Interestingly, unlike STEGO, EQUSS shows different entropy values for different classes.
To understand the relationship between entropy and class, we plot the frequency of each codeword (x-axis) being selected for each codebook (y-axis) in Fig.~\ref{fig:analysis}(b).
We observe that classes containing various objects tend to have high entropy, while those with similar appearances have low entropy.
For example, in classes consisting of various objects such as `Sports' (e.g., surfboard, kite, skis, $\ldots$) and `Indoor' (e.g., toothbrush, vase, clock, $\ldots$), a variety of codewords are activated.
In contrast, classes that consistently appear in similar forms such as `Sky' and `Water' tend to concentrate on specific codewords that are assigned more frequently than others. 
Our analysis verifies that EQUSS indeed reflects the diversity of classes, which aligns with our common intuition.

We conjecture that the class with diverse appearance may confuse the model to learn consistent representation, thus leading to poor performance.
To investigate this, we plot per-class accuracy and its corresponding entropy in Fig.~\ref{fig:analysis}(c).
Note that the classes are sorted in ascending order according to their entropy values. 
As expected, the accuracy tends to decrease as the number of bits needed to represent a class increases.
From these results, we establish the quantitative relationship between entropy and accuracy for the first time.

\begin{figure*}[!t]
    \centering
    \includegraphics[width=0.95\textwidth]{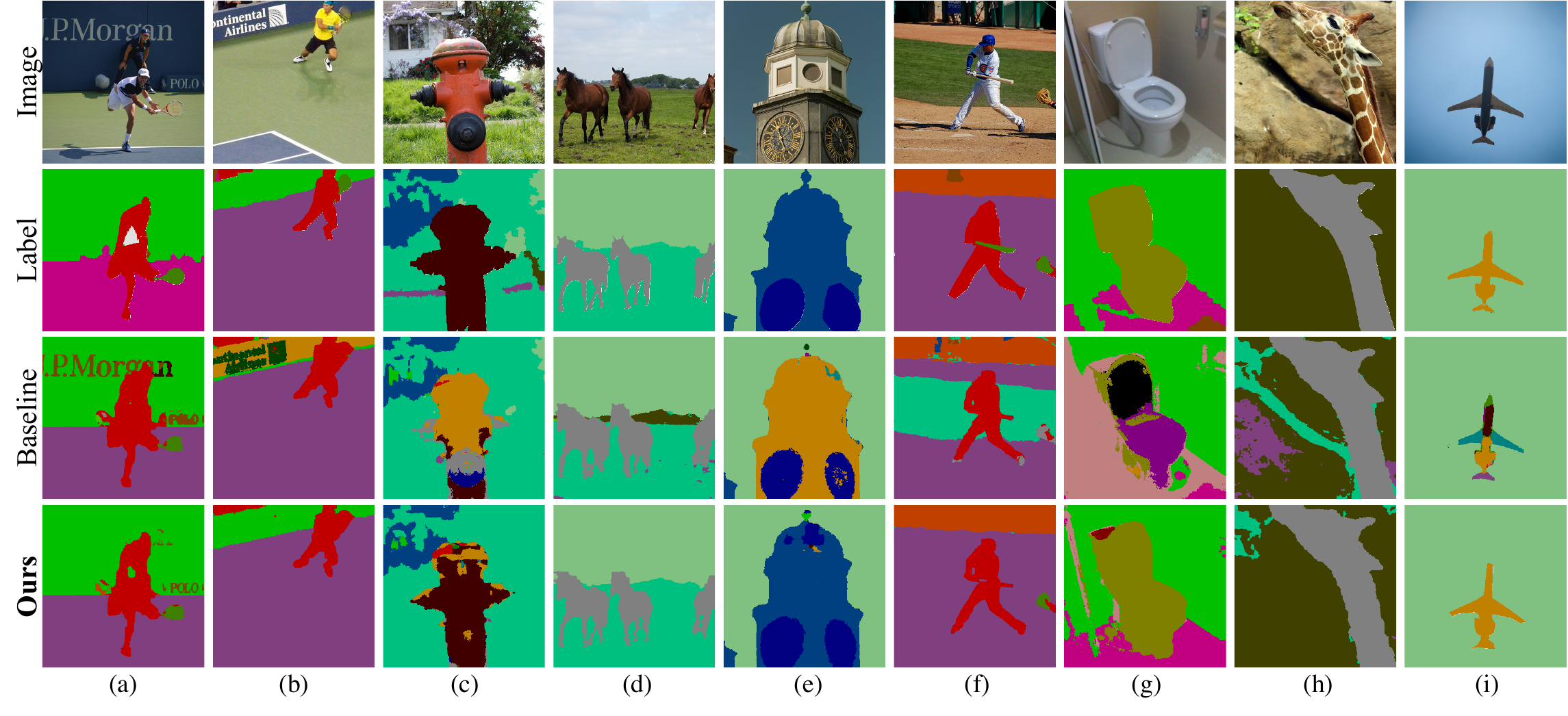}
    \caption{
    Qualitative comparison between the baseline (STEGO) and ours (EQUSS) on \textbf{CocoStuff-27}. 
    }
    \label{fig:comparison}
\end{figure*}
\subsection{Distances Between Classes After PQ}\label{ab:pq}
As illustrated in the blue box of Fig.~\ref{fig:overall}, the combination of selected codewords within the same class would exhibit a strong correlation, whereas those between different classes would have a weak correlation.
To investigate such a correlation, we calculate the distance between codeword combinations of two pixels.
Specifically, we define a codeword combination of $q$ as a sequence of selected codeword indices:
\begin{equation}
    \texttt{combination}(q) = [k^1, k^2, \ldots, k^M]
\end{equation}
where $k^i$ is the $\arg\min$ index of $i$-th subvector (see Eq.~\eqref{eq:argmin}).
Then, we define the distance between two combinations as the sum of the number of different elements as below:
\begin{equation}
    \texttt{distance}(\texttt{c}(q_1), \texttt{c}(q_2)) = \sum_{i=1}^{M} \mathbb{I}(k_1^i \neq k_2^i)
\end{equation}
where $\texttt{c}(q)$ is an abbreviation of $\texttt{combination}(q)$.
Fig.~\ref{fig:analysis}(d) shows the average distance of codeword combinations between classes.
Note that we randomly select 10,000 pixels from every class for distance computation.
We observe that the diagonal of the matrix, corresponding to the distance within the same class, shows much smaller values than other elements of the matrix.
This implies that our quantization head outputs similar results for the same class, which enables clustering to become easier.
\subsection{Ablation Study}\label{ab:ablation}
\paragraph{Component Ablation}
Table~\ref{tab:ablation_cocostuff} shows the effect of feature dimension expansion and product quantization.
As discussed in Sec.~\ref{sec:method:motivation}, naively increasing the feature dimension hurts the performance, specifically resulting in a consistent decrease in linear probing accuracy and mIoU (see M1, M2, and M3).
When PQ is applied for compression, the model achieves higher performance in all cases compared to not using PQ (see M2 \textit{vs.} Q2, M3 \textit{vs.} Q3).
Furthermore, the effectiveness of PQ increases as the feature dimension size increases (see Q2 $<$ Q3).
By combining both feature dimension expansion and product quantization, we can achieve the best result (see Q3).

\begin{figure}[t]
 \begin{center}
 \vspace{-0.8cm}
   \includegraphics[width=1\linewidth]{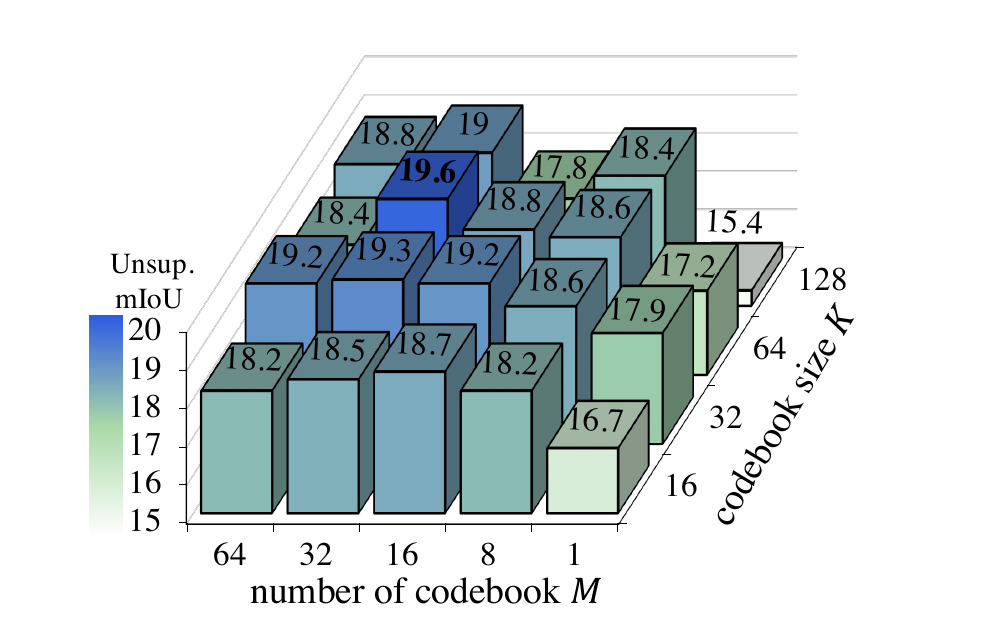}
\end{center}
\vspace{-0.2cm}
    \caption{
    Unsupervised mIoU performance regarding the number and size of the codebook on the \textbf{Cityscapes}.
    Note that the dimension of the feature is fixed to 512.
    }
    \label{fig:pq}
    \vspace{-0.5cm}
\end{figure}

\input{table/ablation_cocostuff.tex}\paragraph{PQ Ablation}
To study the effect of the number of codebooks (\textit{M}) and the size of the codebook (\textit{K}), we conduct experiments while varying \textit{M} and \textit{K} with fixed feature dimension.
Fig.~\ref{fig:pq} presents unsupervised mIoU performance on Cityscapes.
The results show that the performance generally improves as \textit{M} increases.
Specifically, \textit{M} $>1$ outperforms \textit{M}=1 in all cases.
Note that \textit{M}=1 is identical to VQ.
Furthermore, a larger codebook size (\textit{K}) usually improves performance since it expands the number of cases that a single codebook can represent.

However, excessively large values of \textit{M} or \textit{K} slightly reduce the performance.
When the fixed input dimension is divided into too many small subspaces (i.e., \textit{M}=64), such small dimensions of each subvector would restrict their ability to capture necessary representations.
Additionally, for overly large \textit{K} (i.e., \textit{K}=128), each codeword may not receive sufficient updates due to the limited number of training samples, leading to unstable performance.
Nonetheless, in most cases where $\textit{M}\in\{8,16,32\}$ and $\textit{K}\in\{32,64\}$, the performance is consistently higher than our baseline.


%% file: table/ablation_cocostuff.tex
\begin{table}[t]
\centering
\resizebox{1.0\linewidth}{!}{
\begin{tabular}{ccccccccc}
    \toprule
    & \multicolumn{1}{c}{\multirow{3}{*}{\rotatebox{90}{Dim.}}} & \multicolumn{1}{c}{\multirow{3}{*}{\rotatebox{90}{Quant.}}} & \multicolumn{1}{c}{} & & \\
\multirow{2}{*}{Model} & \multicolumn{1}{c}{} & \multicolumn{1}{c}{}  & \multicolumn{2}{c}{Unsupervised} & \multicolumn{2}{c}{Linear Probe}\\
& \multicolumn{1}{c}{} & \multicolumn{1}{c}{}  & Acc & mIoU & Acc & mIoU\\
\midrule
    M1  & 70  &     &  48.3  & 24.5 & 74.4 & 38.3 \\
    M2  & 384 &     &   48.6   & 24.2 & 67.9 & 31.6 \\
    M3  & 1024 &    &  48.1  & 21.9 & 64.6 & 31.5 \\
    \midrule
    Q2  & 384 &  \cmark & 49.7  & 25.1 & 72.4 & 36.7 \\
    Q3 (\textbf{EQUSS}) & 1024 & \cmark &   \textbf{53.8} & \textbf{25.8} & \textbf{75.2} & \textbf{41.2}\\       
    \bottomrule
\end{tabular}}
    \caption{
     Ablation study on \textbf{CocoStuff-27}.
    }
    \label{tab:ablation_cocostuff}
    \vspace{-0.5cm}
\end{table}

%% file: section/6_conclusion.tex
\section{Conclusion}\label{sec:conclusion}
In this paper, we proposed EQUSS, a novel USS framework that takes both advantages of high-dimensional space for improved clustering and product quantization for effective information compression.
From this design, we achieved state-of-the-art performance on three USS benchmark datasets.
For the first time, we quantified the information capacity of USS features and discovered meaningful analyses such as class-specific tendency and the relationship between entropy and accuracy.
We are delighted to emphasize that our novelty involves new insights and directions within the field of USS, contributing to a new information-theoretic understanding of USS.

%% file: section/7_appendix.tex
\renewcommand*\thetable{\Alph{table}}
\renewcommand*\thefigure{\Alph{figure}}

\section*{Appendix}
\subsection*{Loss Ablation}
Table~\ref{tab:ablation_loss} shows the effect of each component on the overall loss function.
We observe that removing the loss for expansion head, $L_{head}$, results in unstable performance.
Since the expansion head generates the input of the quantization head, removing this loss makes it difficult for the codebook to contain the rich representation of codewords.
On the other hand, $L_{commit}$ and $L_{codebook}$ improve the performance by \textbf{+1.6} and \textbf{+2.1} unsupervised accuracy, respectively.
Finally, CRF post-processing~\cite{crf} slightly improves performance.
\input{table/ablation_loss}

\subsection*{Performance Considering Class Imbalance}
Many popular USS datasets, such as CocoStuff-27 and Cityscapes, exhibit class imbalance where the number of pixels per class varies significantly.
This can cause the model to be biased toward the majority classes and fail to learn the characteristics of minority classes.
In other words, the model may falsely predict most of the pixels to majority classes, even if some of them belong to the minority classes.
However, the current unsupervised accuracy (Acc.) metric is not suitable for capturing and handling this imbalance problem because Acc only counts the number of the correct pixels without considering their classes.

\input{table/class_acc}
\begin{figure}[!t] %
    \centering
    \includegraphics[width=1\linewidth]{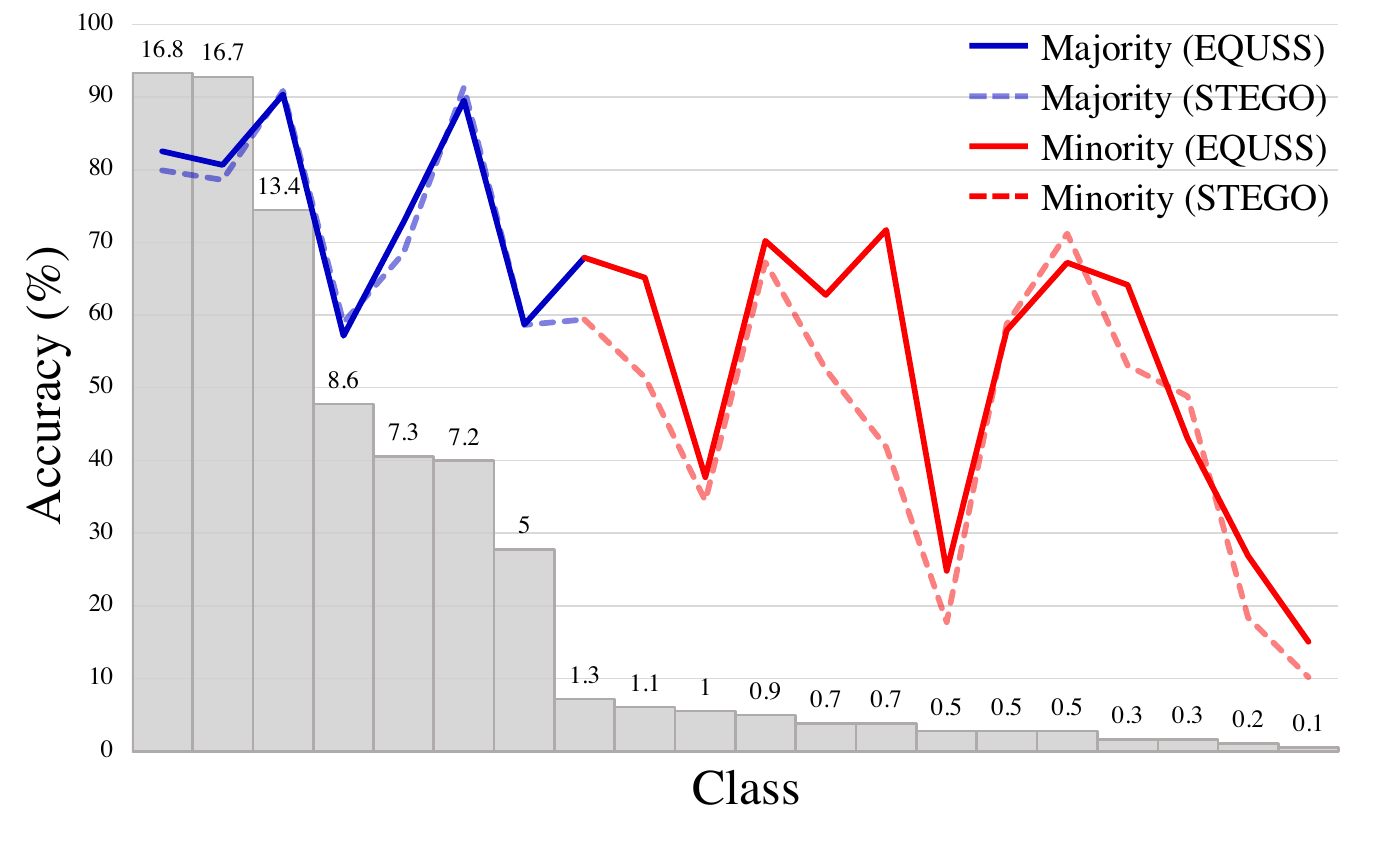}
    \caption{
    Class-wise accuracy of EQUSS (solid) and STEGO (dashed) on \textbf{CocoStuff-27}.
    The classes are sorted in descending order based on their proportion (\%), denoted as numbers on top of each bar.
    The top 7 classes (majority, blue) take more than 5\%, while the bottom 13 classes (minority, red) occupy less than 2\%.
    }
    \vspace{-0.1cm}
    \label{fig:class_acc}
\end{figure}

To address this issue, we suggest measuring \textit{mean class accuracy} (mAcc) for USS metrics.
The mAcc is the average between the pixel accuracy over all classes.
Compared to the conventional Acc metric, mAcc can be an effective metric for representing the overall model performance.
This is because mAcc can consider all classes equally regardless of their class frequency.
Table~\ref{tab:class_acc} compares EQUSS to other recent USS frameworks.
We observe that EQUSS surpasses previous works by a large margin, specifically about \textbf{+2.1} and \textbf{+10.7} unsupervised mAcc gap for the STEGO~\cite{STEGO} and TransFGU~\cite{transFGU}, respectively.
Figure~\ref{fig:class_acc} shows the class-wise accuracy of EQUSS (solid line) and STEGO (dashed line) in detail.
We categorize classes into the majority classes (blue) and minority classes (red) by frequency; each occupies more than 5\% and less than 2\% of the total pixels in the training dataset, respectively.  
From the results, we highlight that EQUSS achieves higher accuracy than STEGO for most of the minority classes.
We believe that expanding the dimension of the feature vector enables fine-grained clustering, which can be helpful for capturing the unique characteristics of minority classes.


\section*{Additional Analysis}\label{sup:sec:discussion}

\subsection*{Higher Information Compression Rate}\label{supp:sec:advantages}
Due to the novel architecture design, EQUSS can sufficiently compress information while enjoying the benefits of high-dimensional features for clustering.
In STEGO, for example, a 384-dim feature vector is compressed into the 70-dim vector and it requires 2,240 bits (32-bit $\times$ 70 elements) to express the compressed feature.
In contrast, our approach casts a feature vector to high-dimensional space and then partitions the expanded feature vector into multiple subvectors and quantizes each subvector.
Specifically, EQUSS expands the feature to a 1024-dim vector and partitions to 64 groups (16-dim for each group) with 256 codewords, each being quantized to 8-bit.
We empirically show that 256 codewords (8-bit) are sufficient for improving performance.
After PQ, the final 1024-dim output only needs 512 bits (8-bit $\times$ 64 groups) for the representation.
In this example, EQUSS could use approximately 15 ($\approx1024/70$) times larger feature dimensions while requiring only a quarter of bits, although the effective number of bits may be smaller.
This aligns with our observation in Fig. 4(a) that EQUSS uses much fewer bits than STEGO to represent each class.

\begin{figure}[t]
    \centering
    \includegraphics[width=1\linewidth]{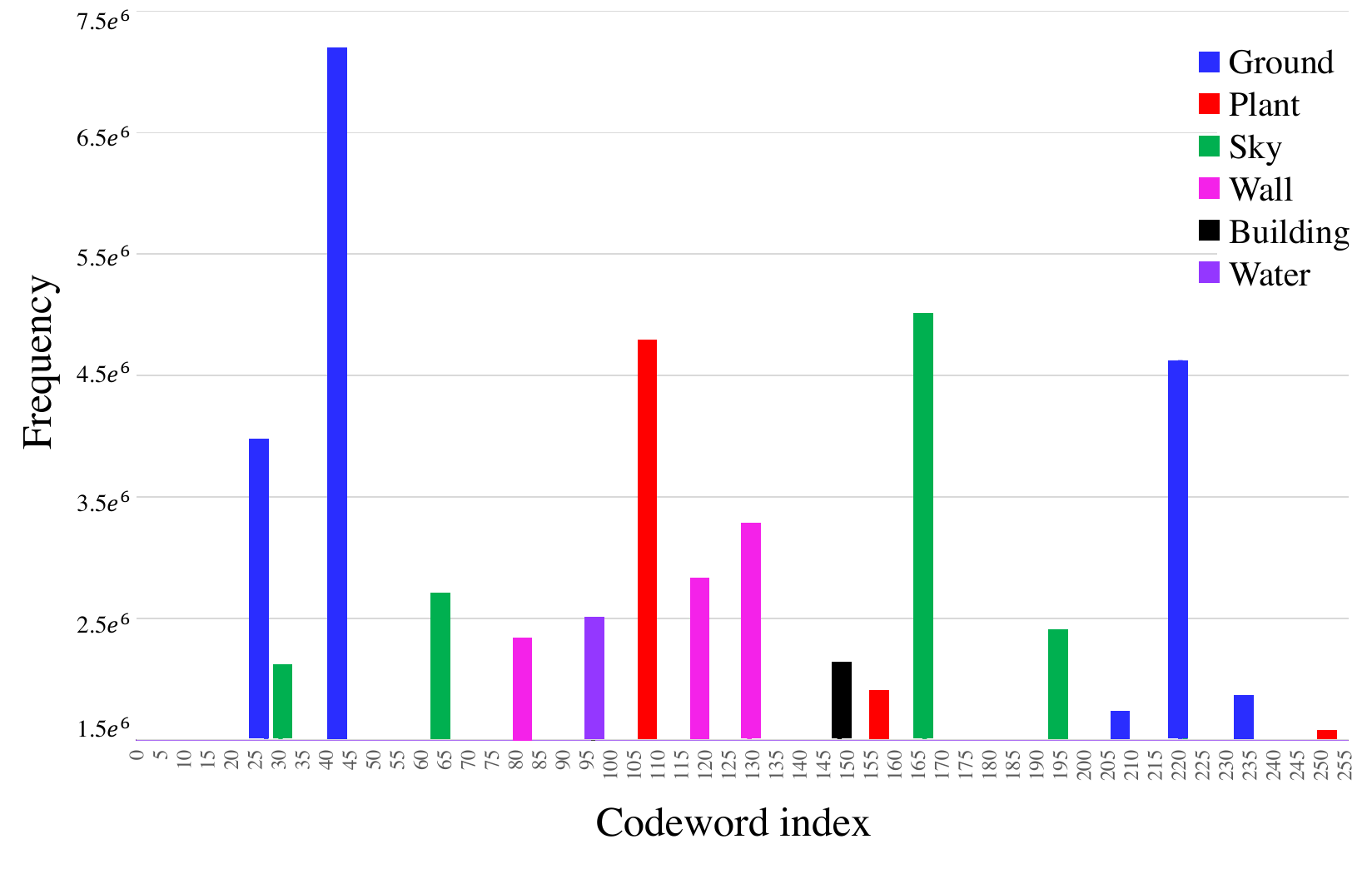}
    \vspace{-0.6cm}
    \caption{
    Histogram that shows the frequency of selected codewords in a single codebook for top 6 majority classes on \textbf{CocoStuff-27}.
    }
    \label{fig:big6}
\end{figure}

\subsection*{Class-specific Codewords}
We expect that certain codewords will show a strong correlation to specific classes, as demonstrated in Fig. 4(b).
To further investigate this class-specific tendency, we visualize the frequency of each codeword in the \textit{first} codebook for the top-6 majority classes in CocoStuff-27.
Fig.~\ref{fig:big6} shows the histogram of each codeword frequency to be selected.
For simplicity, we plot the bar only when the frequency is greater than a given threshold.
The results suggest that certain codewords are particularly activated for specific classes.
In other words, these codewords can be considered unique for each class without being occupied mutually. 
For example, the subvector of \textit{ground} class pixels is often quantized to codewords whose indices are 43, 220, and 25, whereas the subvector of \textit{plant} class pixels is often quantized to codeword number 109. 
We believe that such codewords can perform as class-specific attributes~\footnote{This attribute is similar to the concept of generalized zero-shot learning~\cite{gzsl2}}.
Consequently, some classes have their own representative codeword combination, and their variants do not change much from it (see Fig. 4(d)).

\begin{figure}[t]
    \vspace{-0.5cm}
    \centering
    \includegraphics[width=0.7\linewidth]{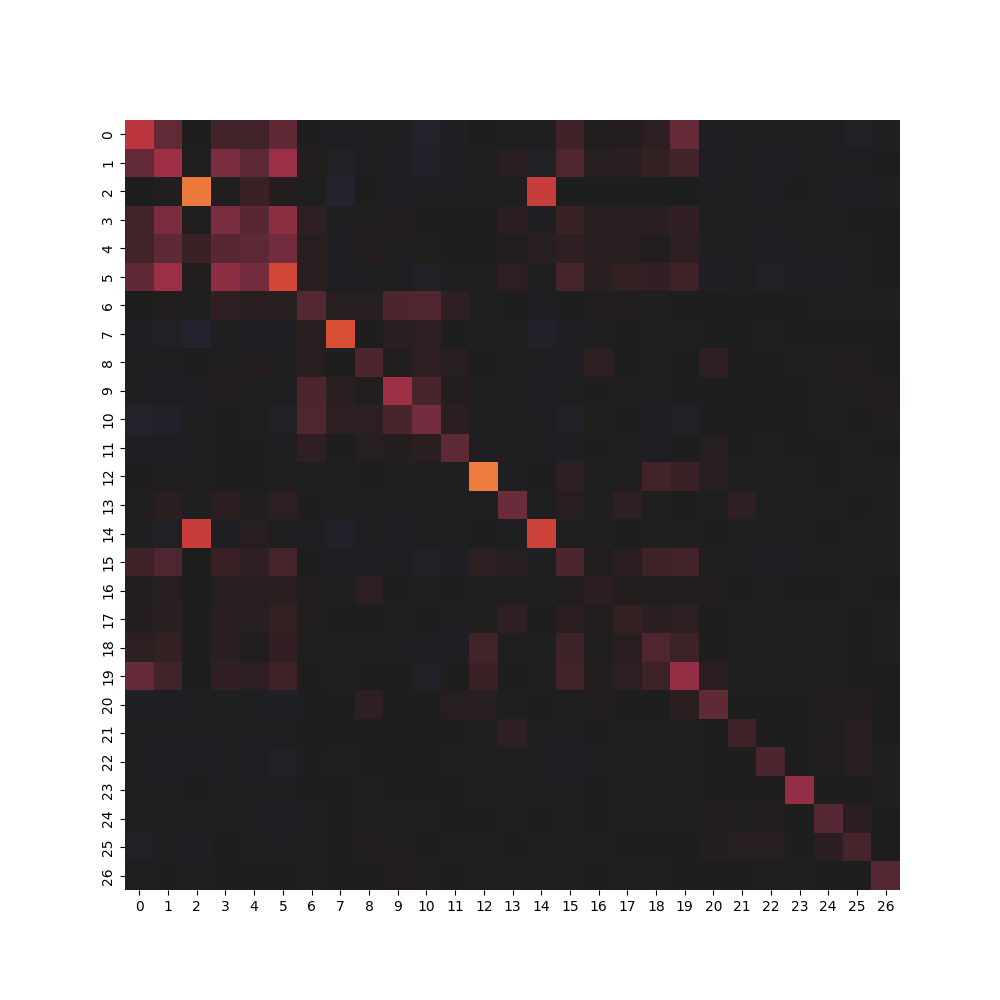}
    \caption{
    The average distance of quantized vectors between two classes. 
    Each axis represents the class index.
    The brighter the color, the closer the distance.
    Best viewed in PDF.
    }
    \label{fig:quantized_matrix}
\end{figure}

\subsection*{Distances Between Quantized Vectors}
We calculate the Euclidean distance between quantized vectors using the Cocostuff-27.
In Fig.~\ref{fig:quantized_matrix}, we plot the average Euclidean distance between two classes.
We see that a large part of the diagonal component is conspicuous, indicating high similarity among quantized vectors belonging to the same class.
In our work, instead of measuring the Euclidean distance between two quantized vectors, we measured the similarity by comparing the number of non-equal codeword indices in two quantized output vectors.
Although the two approaches seem to be different mathematically, the underlying principle of both is more or less similar.

\subsection*{Significance of Performance Improvement}
Our major finding is that EQUSS shows excellent performance for both mIoU and accuracy, but such is not the case for conventional techniques.
For example, while the accuracy difference between EQUSS and TransFGU is a bit marginal, EQUSS significantly surpasses TransFGU in terms of mIoU (25.8 vs. 17.5, see Tab. 1).
Similarly, mIoU scores of EQUSS and STEGO seem to be comparable but EQUSS outperforms SETGO in terms of accuracy by a large margin (53.8 vs. 48.3, see Tab. 1).
We believe that the performance gain of EQUSS is substantial and meaningful.

\section*{Measuring the Information Capacity of Continuous Feature Vector}\label{supp:sec:entropy_stego}
As discussed in Sec. 3.6, it is straightforward to compute the amount of information of features, $J(X)$, for EQUSS.
However, unlike EQUSS, other USS frameworks encode information in the continuous feature vector.
Therefore, we cannot directly apply Eq. (6) for calculating $p^m_i$ in the case of STEGO.
To alleviate this issue, we approximately obtain a discrete distribution of values by histogram.
Specifically, for $m$-th element of the feature vector, we compute the frequency histogram $H^m$ and convert it to the probability.

\setcounter{equation}{9}
We first uniformly divide the min-max range into $B^m$ bins, where the edges of bins are $\{ b^m_0, b^m_1, \ldots, b^m_{B^m} \}$.
Then, the probability of each element $x^m$ to appear within the $i$-th bin is calculated as:
\begin{equation}
    \hat{p}_i^m = \mathbb{E}_X[P(b^m_{i-1} < x^m \le b^m_{i})] \approx \frac{|H^m_{i}|}{|X|} 
\end{equation}
where $|H^m_{i}|$ is the frequency count of histogram corresponds to $i$-th bin.
We can consider this as a uniform quantization from $x^m$ to $ (b^m_{i-1} + b^m_{i}) / 2$, a center point of $i$-th bin.
In doing so, we can formulate a similar equation to Eq. (7) with the approximated probabilities:
\begin{equation}
    \hat{J}{(X)} = \sum_{m=1}^{d_R} \bigg(-\sum_{i=1}^{B^m} \hat{p}_i^m \log_2(\hat{p}_i^m) \bigg). 
    \label{eq:STEGO_entropy}
\end{equation}
$d_R$ is a feature dimension after dimensionality reduction.
Note that the number of bins $B^m$ can vary for each element.
To determine $B^m$, we increase the number of bins while the condition is satisfied: the average quantization error is less than a threshold $\delta^m$.
We set $\delta^m$ to 0.1\% of the total range ($\max(x^m) - \min(x^m)$).
We empirically found that a 0.1\% error rate is sufficient to maintain the original performance of STEGO.


\section*{Detail}\label{supp:sec:details}
\subsection*{Dataset}
We evaluate our model on three standard datasets, CocoStuff-27~\cite{COCO}, Cityscapes~\cite{cityscapes}, and Potsdam-3.
CocoStuff is a large-scale scene-centric dataset with 80 \textit{things} categories and 91 \textit{stuff} categories. 
We evaluate the model on 27 mid-level classes of the dataset hierarchy.
Cityscapes contains a wide variety of urban street scenes from 50 different cities in good weather conditions.  
We utilize 27 classes excluding \textit{void} group for the evaluation.
Potsdam-3 is a collection of satellite images obtained by extracting patches of $200\times200$ from raw images of high resolution.
We evaluate our model on the 3-label variants including roads, vegetation, and buildings.

Following prior art~\cite{STEGO}, we apply a five-crop transformation on training samples to better handle small objects.
We resize and center-crop the training and validation samples to (224, 224) and (320, 320) resolutions, respectively.
For evaluation, we measure mean intersection over union (mIoU) and per-pixel accuracy.

\input{table/hyper.tex}
\subsection*{Evaluation Protocol}
\paragraph{Linear Probing}
We train a single linear projection layer for the supervised classification task by utilizing ground truth labels.
The main purpose of linear probing is to evaluate how much class-relevant information is contained in the final output features.

\paragraph{Unsupervised Clustering}
Unlike linear probing, we do not access any labels and only cluster the output features in an unsupervised manner.
Then, we use a Hungarian matching algorithm to compute the optimal assignment from our unlabeled clusters to ground truth labels.

\subsection*{Implementation Detail}
We train our model with a batch size of 16 on a single NVIDIA A5000 24GB GPU.
Following previous work~\cite{STEGO,transFGU}, we employ DINO~\cite{DINO} pre-trained on ImageNet-22K~\cite{imagenet} as a backbone. 
For a fair comparison~\cite{STEGO,transFGU}, we keep the backbone frozen.
We initialize each codebook in the quantization head $\mathcal{Q}$ by Xavier uniform initialization~\cite{xavier_uni}.
The expansion head $\mathcal{E}$ is a feed-forward network having a residual path constructed by a two-layer MLP. 
For training, we employ Adam optimizer~\cite{adam} with a learning rate of 3e-4.
For linear probing and unsupervised clustering, we use Adam optimizer with a learning rate of 3e-3.
We set $\lambda_1=1.0$ and $\lambda_2=0.25$ in Eq. (5) for the experiments.

The output dimension of the expansion head is 1024, and the DINO~\cite{DINO} backbone feature dimensions of ViT-S and ViT-B architecture~\cite{vit} are 384 and 768, respectively.
The expansion head has two branches, \texttt{Conv-ReLU-Conv} and \texttt{Conv}, that process the backbone feature separately. 
The output of the expansion head is the addition of two outputs from branches.
\texttt{Conv} layers employ 1x1 kernel size with 1x1 stride.
Feature dimension expansion to the desired dimension occurs at the first convolutional layer of each branch.

\subsection*{Hyperparameter}
Table~\ref{tab:hyper} shows the hyperparameters and architecture (PQ) details used in our experiments.
Note that $\lambda$ and $b$ control $L_{head}$, which is originated from STEGO~\cite{STEGO}.

\section*{Additional Qualitative Result}\label{supp:sec:qualitative}
\begin{figure*}[!t]
    \centering
    \includegraphics[width=1.0\textwidth]{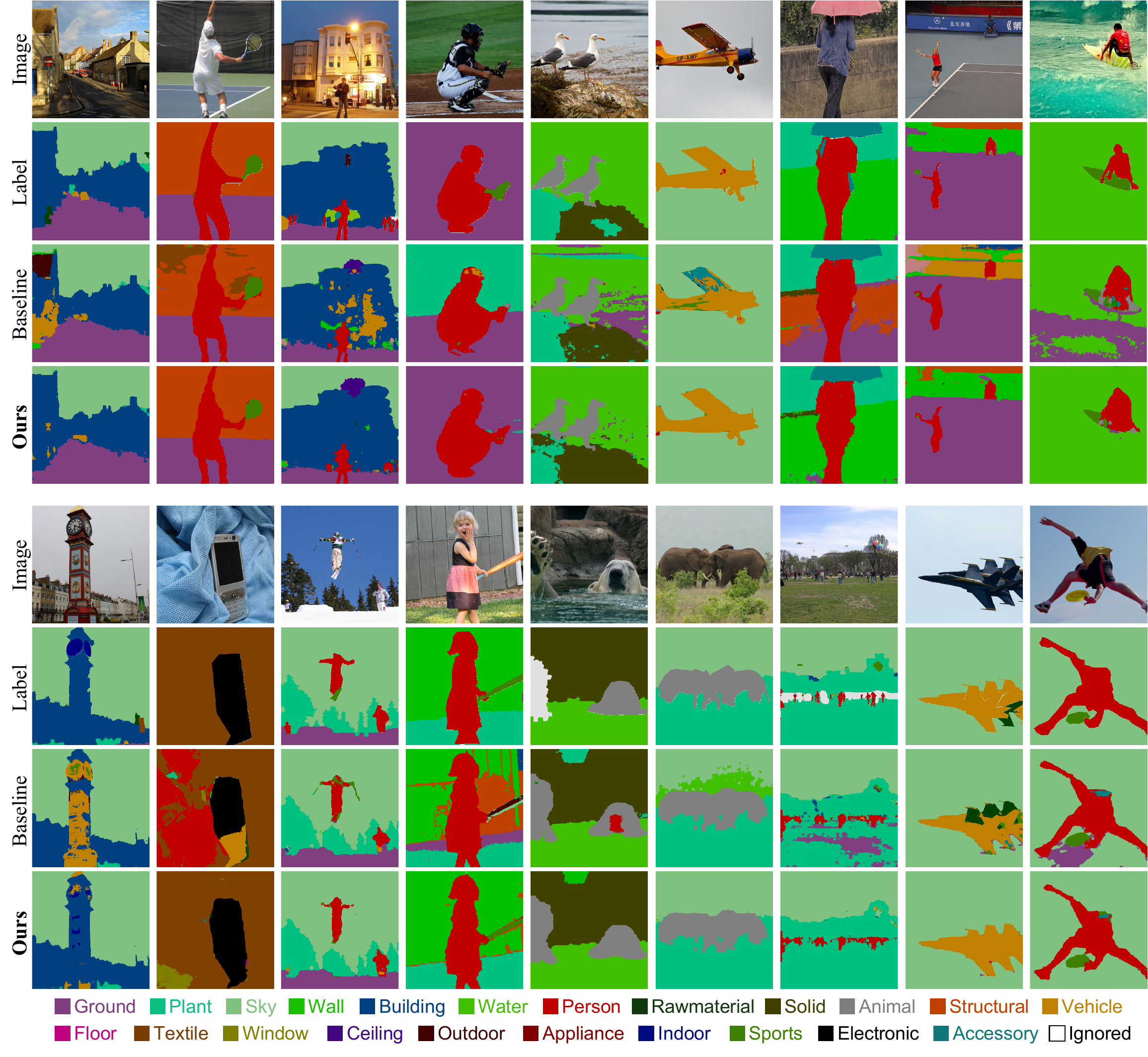}
    \caption{
    Qualitative comparison between the baseline (STEGO) and ours (EQUSS) on \textbf{CocoStuff-27}. 
    }
    \label{fig:supp_visual}
\end{figure*}

Figure~\ref{fig:supp_visual} shows the additional qualitative comparison between the baseline~\cite{STEGO} and ours (EQUSS).

\section*{Limitation}
\textit{Limitations} of our methods include: (1) since our work is the first to focus on the advantage of feature dimension expansion and quantization, we simply implement grouping by dividing the features into the same channel size.
Thus, we believe that dynamically adjusting the number of channels for each group can further improve performance and leave it as a future work;
(2) currently, the interpretability between the codewords and human perceivable attributes is quite low.
Thus, training the models to capture intuitively explainable human-like attributes would be meaningful research.

\section*{Discussion}
We would like to emphasize that our main contribution is to address the fundamental weakness of bottleneck architecture in previous studies and propose a novel framework improving both clustering and information compression.
Additionally, we highlight that our work is the first to analyze USS from the entropy aspect.
Specifically, we have investigated the relationship between the USS performance and entropy and verified that the entropy effectively reflects the diversity of classes (see Sec. 5.1).
We believe that our research can contribute to improving the understanding of USS from an information-theoretic perspective.

%% file: table/ablation_loss.tex
\setlength{\tabcolsep}{9pt}
\begin{table}[h]
    \centering
    \resizebox{1.0\linewidth}{!}{
    \begin{tabular}{cccc|cc}
    \toprule
           &     &        &   &  \multicolumn{2}{c}{Unsupervised}\\
    $\textit{L}_{head}$ & $\textit{L}_{commit}$ & $\textit{L}_{codebook}$   & CRF & Acc. & mIoU  \\ 
    \midrule
            & \cmark &   \cmark      &     & 16.8 & - \\
    \cmark &        &  \cmark       &  & 50.7 & 24.1\\
    \cmark & \cmark &         & & 50.2& 24.3  \\
    \cmark & \cmark & \cmark & & 52.3 & 25.3 \\
    \midrule
    \cmark & \cmark & \cmark & \cmark & \textbf{53.8} & \textbf{25.8} \\
    \bottomrule
    \end{tabular}
    }
    \caption{
     Ablation study on \textbf{CocoStuff-27}.
    }
    \label{tab:ablation_loss}
\end{table}
\setlength{\tabcolsep}{6pt}

%% file: table/class_acc.tex
\setlength{\tabcolsep}{11pt}
\begin{table}[t]
    \centering
    \resizebox{1.0\linewidth}{!}{
    \begin{tabular}{lcc}
    \toprule
        Model & Unsup. mAcc. & Linear. mAcc. \\ 
        \midrule
        PiCIE & 25.1  & 27.0 \\ 
        TransFGU & 26.7  & 31.7 \\ 
        STEGO & 35.3  & 59.7 \\ 
        \midrule
        \textbf{EQUSS} (Ours) & \textbf{37.4} & \textbf{61.0}  \\ 
        \bottomrule
    \end{tabular}
    }
    \vspace{0.1cm}
    \caption{Mean class accuracy (mAcc) on \textbf{CocoStuff-27}. 
    The scores for other models are measured using the published pre-trained weights.
    }
    \vspace{-0.1cm}
    \label{tab:class_acc}
\end{table}
\setlength{\tabcolsep}{6pt}

%% file: table/hyper.tex
\begin{table}[t]
    \centering
    \resizebox{1.0\linewidth}{!}{
    \begin{tabular}{ccccc}
    \toprule
        & Cocostuff-27 & & Cityscapes & Potsdam-3 \\ 
        \midrule
        \multicolumn{4}{l}{Codebook configuration.} \\
        \midrule
        \#Codewords  & 256 & & 32 & 16 \\ 
        \#Codebooks  & 64 & & 32 & 64 \\ 
        Codeword dim. & 16 & & 32 & 16 \\
        \midrule
        \multicolumn{4}{l}{Loss hyperparameters for $L_{\text{head}}$.} \\
        \midrule
        $\lambda_{rand}$ & 0.63 & & 0.95 & 0.63 \\ 
        $\lambda_{knn}$ & 0.25 & & 0.43 & 0.25 \\ 
        $\lambda_{self}$ & 0.67 & & 1 & 0.67 \\ 
        $b_{rand}$ & 0.66 & &0.31 & 0.26 \\ 
        $b_{knn}$ & 0.02 & & 0.22 & 0.12 \\ 
        $b_{self}$ & 0.08 & & 0.36 & 0.21 \\ 
    \bottomrule
    \end{tabular}
    }
    \caption{
    Hyperparameters for different datasets.
    }
    \label{tab:hyper}
\end{table}

%% file: section/8_acknowledgment.tex
\section*{Acknowledgments}\label{sec:acknowledgments}
This work was supported by the National Research Foundation of Korea (NRF) grant funded by the Korea government (MSIT) (2022M3C1A3099336) and National Research Foundation of Korea (NRF) grant funded by the Korea government (MSIT) (RS-2023-00208985).